\pdfoutput=1

\documentclass[11pt]{article}

\usepackage[]{ACL2023}

\usepackage{times}
\usepackage{latexsym}
\usepackage{graphicx}
\usepackage{algorithm, algorithmic}
\usepackage{amsmath}
\usepackage{booktabs}
\usepackage{tabularx}
\usepackage{amssymb} 
\usepackage{array} 
\usepackage{multirow} 
\newcommand{\best}[1]{\textcolor[RGB]{205,83,76}{#1}}
\newcommand{\secondb}[1]{\textcolor[RGB]{0,115,194}{#1}}

\usepackage[T1]{fontenc}

\usepackage[utf8]{inputenc}

\usepackage{microtype}

\usepackage{inconsolata}

\title{Knowledge-Infused Legal Wisdom: Navigating LLM Consultation through the Lens of Diagnostics and Positive-Unlabeled Reinforcement Learning}

\author{Yang Wu\textsuperscript{$\clubsuit$}\thanks{\hspace{1mm} Corresponding author} \quad Chenghao Wang\textsuperscript{$\heartsuit$} \quad Ece Gumusel\textsuperscript{$\diamondsuit$} \quad Xiaozhong Liu\textsuperscript{$\clubsuit$}  \\
\textsuperscript{$\clubsuit$}Worcester Polytechnic Institute \quad \textsuperscript{$\heartsuit$}Peking University \quad \textsuperscript{$\diamondsuit$}{Indiana University at Bloomington}
\\
\texttt{\{ywu19, xliu14\}@wpi.edu \quad chenghao.wch@gmail.com \quad egumusel@iu.edu}
}

\begin{document}
\maketitle

\begin{abstract}
The integration of generative Large Language Models (LLMs) into various applications, including the legal domain, has been accelerated by their expansive and versatile nature. However, when facing a legal case, users without a legal background often struggle to formulate professional queries and may inadvertently overlook critical legal factors when presenting their case narrative to LLMs. To address this issue, we propose the Diagnostic Legal Large Language Model (D3LM), which utilizes adaptive lawyer-like diagnostic questions to collect additional case information and then provides high-quality feedback. D3LM incorporates an innovative graph-based Positive-Unlabeled Reinforcement Learning (PURL) algorithm, enabling the generation of critical questions and enhancing user-LLM interactions. Moreover, an integrated LLM-based stopping criterion facilitates precise Court Views Generation (CVG). Our research also introduces a new English-language CVG dataset\footnote{https://github.com/YANGWU001/\text{US\_CVG\_dataset}.git} based on the US case law database, enriching the realm of LLM research and deployment with a vital dimension. D3LM surpasses classical LLMs by delivering outstanding performance and a remarkable user experience in the legal domain.

\end{abstract}

\section{Introduction}

Recent years have witnessed a substantial increase in the demand for legal services (\citealp{gans2017demand,purba2019demand}). Despite this growing need, legal resources remain scarce, especially for individuals with modest means \citep{hadfield2010higher}. This scarcity has led to a significant rise in the cost of legal assistance \citep{brescia2014embracing}, making it difficult for ordinary people to afford the necessary legal support. Furthermore, this situation exacerbates the inherent inequities in legal proceedings between the affluent and the underprivileged. Affluent individuals often benefit from dedicated legal teams to represent them in disputes, whereas the economically disadvantaged lack similar access to legal aid \citep{horwitz20204}. This discrepancy not only hinders the ability of the poor to seek justice but also perpetuates a system of latent unfairness in legal cases across socio-economic divides. \\
\begin{figure*}[t]
  \centering
  \includegraphics[width=1.0\linewidth]{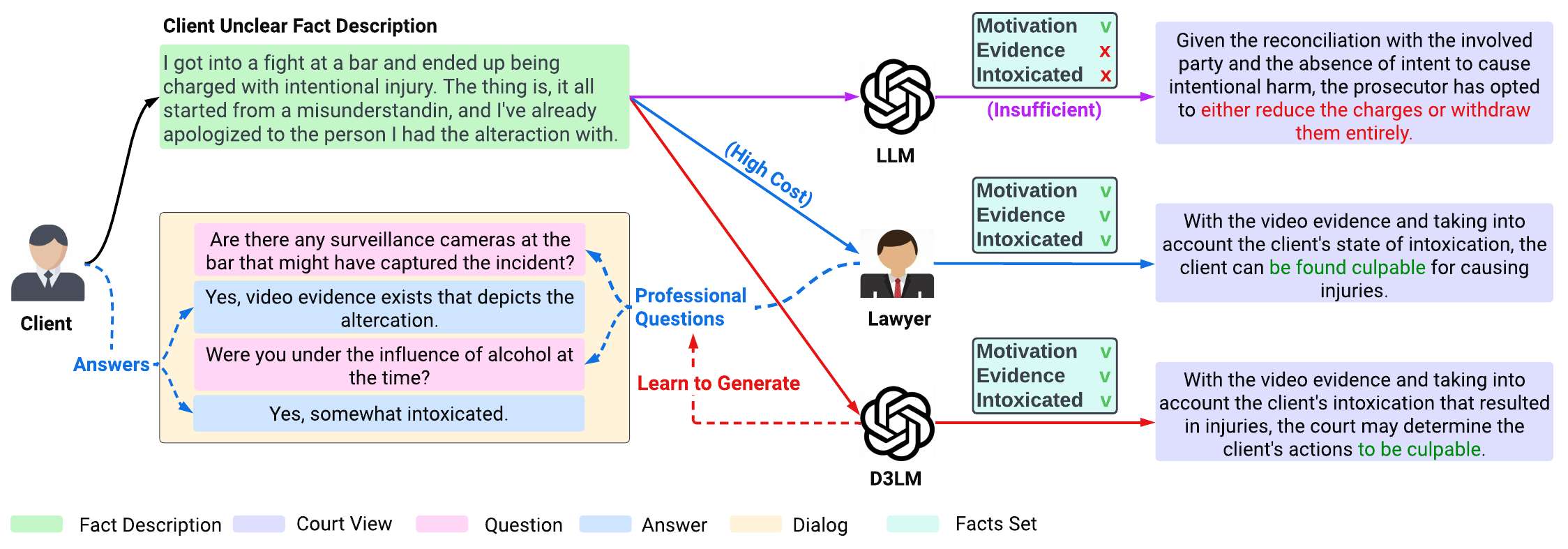}
  \caption{Comparison of legal service methodologies, highlighting traditional LLMs, lawyer consultations, and the D3LM model. D3LM innovatively generates professional questions, mirroring the actions of a lawyer, to improve legal outcome accuracy without high costs, demonstrating a cost-effective, precise approach to legal assistance.}
  \label{toy_example}
\end{figure*}
With the advent of LLMs, ordinary individuals have gained increased access to legal resources and services. However, the current mode of interaction between users and LLMs predominantly remains user-led (\citealp{achiam2023gpt,touvron2023llama}), where users input queries and receive responses from LLMs. This interaction mode presents significant limitations, particularly as many users lack a legal background \citep{zamfirescu2023johnny}. Their queries often lack the essential factual details about their legal issues, which in turn, hampers the ability of LLMs to provide effective assistance. Recognizing this challenge, we have developed D3LM, a novel framework designed to act as a legal assistant, as illustrated in Fig. \ref{toy_example}. In this toy example, we introduce the D3LM model through a real-life scenario where a client, facing legal charges from a minor altercation, seeks assistance. Initially, the client provides a vague description, prompting a comparison among three different approaches. Traditional LLMs offer a broad perspective of the court's view, but often overlook critical details due to insufficient information. In contrast, consultations with a lawyer provide a thorough investigation through targeted questions, albeit at a significant cost. D3LM distinguishes itself by autonomously generating specific, case-relevant questions, thereby gaining a deeper understanding of the situation and enhancing the accuracy of legal outcome predictions, all without incurring the high expenses associated with legal counsel. By engaging users in a dialogue to extract detailed information about their legal predicaments, D3LM mirrors the proactive approach of human lawyers in gathering pertinent data to better assist their clients. This innovative engagement method ensures D3LM delivers highly accurate and cost-effective legal insights, demonstrating its potential to  transform the provision of legal services.

Moreover, we introduce a novel approach that leverages LLMs to efficiently process and analyze extensive U.S. legal cases. Utilizing LLMs, we generate structured fact-rule graphs and the 'Issue, Rule, Analysis, Conclusion' (IRAC\footnote{The detailed explanation of IRAC is in Appendix \ref{prompts} }) \citep{gensler1985irac} to transform complex case narratives into concise representations. Our method integrates these graphs into a comprehensive knowledge base, designed to identify and fill in missing factual information in legal scenarios. By employing a reinforcement learning approach, we refine the selection of relevant fact nodes, enhancing the model's ability to guide users through legal inquiries effectively. This system is rigorously evaluated to ensure its accuracy and practicality in real-world legal settings, showcasing its potential to innovate legal case analysis.

Our contributions in this paper are threefold and can be summarized as follows:
\begin{itemize}
\item Novel User-LLM Interaction Paradigm: We introduce a new user-LLM conversion framework in legal contexts by shifting from user-led enquiries to LLM-navigated diagnostics. Our approach actively seeks detailed case information from user through targeted questioning, resulting in unparalleled accuracy and efficacy in legal advice.

\item Pioneering Positive-Unlabeled Reinforcement Learning (PURL) Algorithm: Leveraging the fusion of LLM and legal graph, PURL identifies crucial factors dynamically for adaptive question generation, which enabling LLM's information solicitation capabilities in legal contexts.

\item Introducing a new English legal dataset: a new focus on US legal cases. Departing from prior studies relying on Chinese data, this dataset establishes a new benchmark for legal AI experiments in the US case law system.
\end{itemize}




\section{Related Work}

\subsection{Legal Assistant}

Recent advancements in computational methods have significantly impacted judicial decision-making, leveraging the text-centric nature of legal data for natural language processing (NLP) applications. Early research by \citet{lin2012exploiting}, \citet{zhong2018legal}, and \citet{hu2018few} treated charge prediction as a classification problem within legal judgments, laying the groundwork for NLP in legal contexts. This foundation expanded to include case retrieval \citep{chen2013text} and law article recommendation \citep{chen2022few}, enhancing NLP's scope in legal informatics. The interpretability of legal decisions, explored by \citet{ye2018interpretable} and \citep{wu2020biased}, introduced court view generation (CVG) as a means to predict judicial outcomes based on case descriptions, emphasizing the need for nuanced legal text understanding and generation. The introduction of LLMs like Lawformer \citep{xiao2021lawformer} and Chatlaw \citep{cui2023chatlaw} further advanced this field, demonstrating LLMs' potential in generating accurate legal texts. Unlike previous models, our D3LM approach mimics a lawyer's consultative strategy, engaging in dynamic interaction through targeted questioning to gather comprehensive case details, marking a significant evolution in the application of AI for legal informatics.

\subsection{Enhanced Learning through PU Learning and Reinforcement Learning}
In the landscape of computational learning, Positive-Unlabeled (PU) Learning has emerged as a key strategy for binary classification, leveraging a mix of a small number of positive instances and a large volume of unlabeled data. This methodology has been applied in diverse areas including chronic disease prediction (\citealp{chen2020self,wu2023community}), fake news detection \citep{liu2020fned}, and recommendation systems \citep{zhou2021pure}, utilizing advanced techniques like uPU and nnPU \citep{kiryo2017positive} alongside unbiased risk estimators \citep{du2015convex} to mitigate data bias. Innovations such as predictive adversarial learning (\citealp{hu2021predictive, wei2018positive}) and Dist-PU \citep{zhao2022dist} further refine this approach by enhancing data generation and label distribution consistency. Additionally, the adaptation of reinforcement learning, particularly bandit learning, has introduced novel training signals and strategies, including context-free (\citealp{chapelle2011empirical, auer2002finite}) and contextual bandits (\citealp{chu2011contextual, zhou2020neural}), to improve model outcomes. Our work introduces the PURL framework, integrating the advanced capabilities of LLMs with domain-specific PU models to achieve decision-making precision tailored to unique application contexts, signifying a notable advancement in computational learning methodologies.

\subsection{Large Model Dataset Creation}

In the NLP field, LLMs have transformed dataset generation, producing extensive datasets that capture diverse linguistic phenomena, enhancing versatility for various NLP tasks. Examples include WANLI \cite{liu2022wanli}, utilizing GPT-3 for generation and dataset cartography \cite{swayamdipta2020dataset} for refining NLI examples, and DISCO \cite{chen2023disco}, which creates counterfactual examples through local perturbations. However, deploying LLMs in legal domains, especially for U.S. cases, faces challenges like narrative length and complexity, which exceed LLMs' token limits, hindering dataset creation and model training \cite{jin2024llm}. Our approach leverages LLMs to streamline U.S. legal narratives, facilitating manageable CVG dataset creation and bridging English-language resource gaps with legal expert collaboration. This method addresses English legal analysis dataset scarcity, adding significant resources to the domain.

\section{Methodology}
In this section, we begin by introducing the definition of the CVG problem in Section 3.1. We then proceed to present the overall D3LM framework in Fig. \ref{workflow} in Section 3.2. Finally, we demonstrate the PURL algorithm in Alg. \ref{purl} in detail in Section 3.3. To facilitate understanding and clarity, we provide a summary of the notations used in this paper in Tab. \ref{notations}.

\begin{table}[h]
\centering
\small
\caption{Notations.}
\label{tab:notations}
\footnotesize 
\setlength{\extrarowheight}{2pt}
\begin{tabular}{|m{0.2\columnwidth}|m{0.7\columnwidth}|} 
\hline
\textbf{Notations} & \textbf{Descriptions} \\ 
\hline
\( D \) & Cases fact descriptions set. \\
\hline
\( C \) & Cases court views set . \\
\hline
\( \overline{D} \) & Reconstructed fact descriptions set without masked facts. \\
\hline
\( Q \) & Generated questions based on golden court views set. \\
\hline
\( F \) & Candidate fact nodes set. \\
\hline
\( \hat{G} \) & Fact-rule graphs set for cases. \\
\hline
\( G' \) & N-hop graphs set of masked fact-rule graphs. \\
\hline
\( \overline{G} \) & Masked fact-rule graphs set. \\
\hline
\( G \) & Merged fact-rule graph. \\
\hline
\end{tabular}
\label{notations}
\end{table}
\subsection{Problem Definition}
In this study, we address the challenge of generating court views in US criminal cases. Our focus is to formalize this problem through a clear delineation of fact descriptions and court views.\\
\textbf{Fact Description (D)} is the comprehensive narrative of a given legal case, encompassing all pertinent facts and events. We represent fact descriptions as $D\ =\ \left(D_1,\ D_2,...,\ D_{\left|D\right|}\right)$, where each $D_i$ corresponds to the fact description of a unique case.\\
\textbf{Court Views (C)} represent the judicial rationale underlying the court's verdict for a case. We define court views as $C\ =\ \left(C_1,\ C_2,...,\ C_{\left|C\right|}\right)$. The primary task in court views generation is to accurately derive $C$ given $D$. 
\subsection{D3LM Framework}

\begin{figure}[t]
  \centering
  \includegraphics[width=\linewidth]{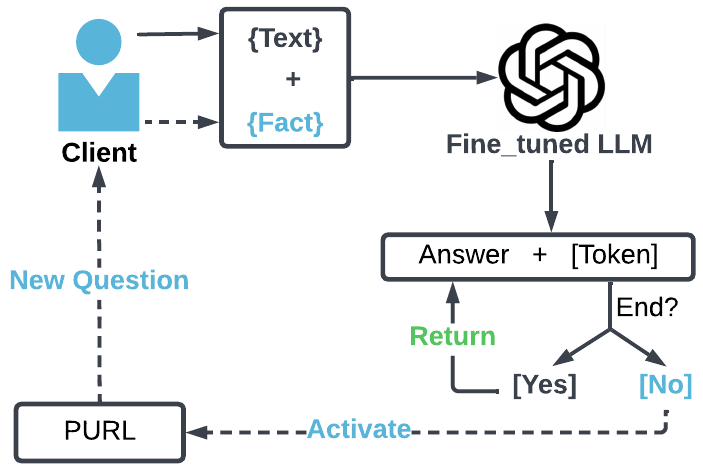}
  \caption{D3LM Model Framework Overview: Illustrates D3LM's engagement through context-driven questions, guided by the PURL algorithm from continuous and historical dialogues. Aims to collect comprehensive case details until a fine-tuned LLM token signals adequate information acquisition.}

  \label{workflow}
\end{figure}

We elaborate on the D3LM framework as illustrated in Fig. \ref{workflow}. Our framework activates upon receiving a legal query from a client, which is then processed by an advanced Language Model (LLM) to formulate an initial court opinion. However, this opinion is not immediately disclosed; it is contingent on the following procedure. Leveraging the methodology proposed by \citet{schick2023toolformer}, we deploy a specialized token to evaluate the completeness of case information provided. This binary token '$Yes$' or '$No$' guides the subsequent actions: a '$No$' activates our PURL algorithm to craft new, case-specific question for client to gather more case information, whereas a '$Yes$' moves the process toward the conclusion and delivery of the court opinion. This dynamic cycle of evaluation and information enhancement continues until a '$Yes$' token signifies the attainment of a fully informed court opinion.


Utilizing a Case-View Generation (CVG) dataset comprising a set of case descriptions $D\ =\ \left(D_1,\ D_2,...,\ D_{\left|D\right|}\right)$ and corresponding court opinions $C\ =\ \left(C_1,\ C_2,...,\ C_{\left|C\right|}\right)$, for each pair ( $D_i$, $C_i$), we randomly mask certain facts from both $D_i$ and $C_i$ to derive $\overline{D_i}$ and $\overline{C_i}$. Subsequently, we append the special tokens $e_i$ at the end of each court view, as:
\begin{equation}
    \begin{aligned}
        e = <\text{T}>\ \rightarrow \ token\ <\text{/ T}>
    \label{token}
    \end{aligned}
\end{equation}
where "$<\text{T}>$", "$\rightarrow$", and "$<\text{/ T}>$" are fixed special tokens\footnote{In practice, we use the token sequences "[", "]" and "->" to represent "$<\text{T}>$", "$<\text{/ T}>$" and "$\rightarrow$"}."$token$" is "$Yes$" for original $C_i$ and "$No$" for masked $\overline{C_i}$. This augmented dataset serves to fine-tune the LLM. During the inference phase, regular decoding proceeds until the fine-tuned LLM outputs the "$\rightarrow$", which dictates the decision to either continue or conclude the information gathering process. By prompting users to provide detailed responses through case-specific questions, the D3LM framework iteratively refines the court view, ensuring that the final advice is accurate and tailored to the client's needs. This innovative method enhances the model's ability to collect pertinent information, leading to more precise legal guidance.

\begin{figure*}[t]
  \centering
  \includegraphics[width=1.0\linewidth]{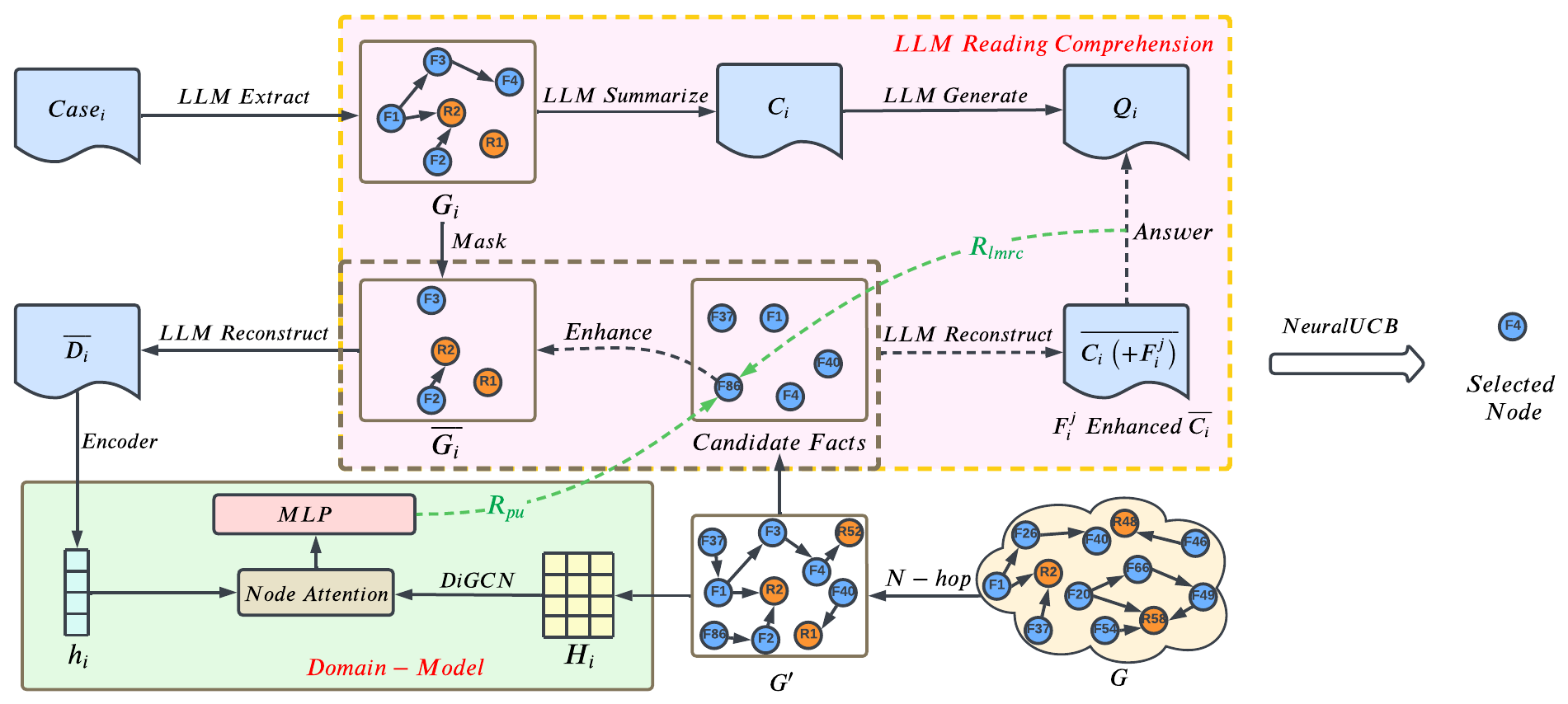}
  \caption{ Illustrative Representation of the PURL Network in Action. This diagram showcases the PURL algorithm's training process using $Case_i$ as an example. It visually delineates the sequential steps of extracting, summarizing, and reconstructing case facts, followed by question generation. Specifically, the collaboration of the LLM with the domain-specific PU model facilitates the final adaptive node selection through NeuralUCB.}
  \label{PURL}
\end{figure*}

\subsection{Positive-unlabeled Reinforcement Learning (PURL) Question Generation}
In this section, we showcase our PURL framework for next question generation through Fig. \ref{PURL} and detail its training process in Alg. \ref{purl}. 


As illustrated in Fig.~\ref{PURL}, for each extensive legal caselaw document, denoted as \(Case_i\), we utilize an LLM to extract a fact-rule graph \(G_i\) representing the case\footnote{The detailed fact-rule graph generation prompt is in Appendix \ref{prompts} }. Concurrently, we apply the legal IRAC strategy to summarize the case's fact description \(D_i\) and court views \(C_i\). Subsequently, the LLM generates a set of questions \(Q_i\) based on the text of \(C_i\). For model training, we randomly mask some fact nodes in \(G_i\) to create a masked graph \(\overline{G_i}\), and then use the LLM to reconstruct the fact description and corresponding court view as \(\overline{D_i}\) and \(\overline{C_i}\), respectively. Based on the nodes in \(\overline{G_i}\), we extract an N-hop subgraph \(G'\) from the knowledge graph \(G\), which is a merged graph comprising \(\left\{G_1, G_2, \ldots, G_{\left|G\right|}\right\}\). We then define the candidate fact set as \(F_i = \left\{v \ | v\in V_{G'_i}\ -\ V_{\overline{G_i}}\right\}\), where $V_{G'_i}$, $V_{\overline{G_i}}$ represent the nodes in $G'_i$ and $\overline{G_i}$, respectively. The task involves selecting the most relevant node from \(F_i\) for the masked case description \(\overline{D_i}\). Following node selection, we use a prompt-based approach to transform the selected node into a case-specific question to client. The detailed process for CVG data generation is formalized in Alg.~\ref{cvg}.

\subsubsection{Domain Positive-Unlabeled Model Training}


Given a masked case description \(\overline{D_i}\) and its corresponding subgraph \(G'\), we initialize the text embedding \(h_i\) using RoBERTa \citep{liu2019roberta} and randomly initialize the node embeddings as a matrix \(H_i \in \mathbb{R}^{n \times d}\), where \(n\) denotes the number of nodes and \(d\) denotes the embedding dimension. After processing the node embeddings through several DiGCN layers \citep{tong2020digraph}, a node attention\footnote{The detailed information of node attention design is in Appendix \ref{node_atta} } network is applied to capture semantic relationships between case description text and candidate fact nodes. Subsequently, a Multilayer Perceptron (MLP) is utilized to compute the probability scores.

For each $\{h_i, H_i^j\}$ pair, we designate the set of masked nodes as positive instances and the remaining nodes as unlabeled instances. During the training process, we employ the nnPU \citep{kiryo2017positive} risk estimator, which is defined as follows:

\begin{equation}
\small
    \begin{aligned}
        {\text{Risk}}_{pu}(g) = &\frac{\pi_p}{n_p} { \sum\limits_{i=1}^{n_p} } l(g(x_i^{p}),+1) + \\
    &max(0,\frac{1}{n_u} { \sum\limits_{i=1}^{n_u} } l(g(x_i^{u}),-1) - \\
    &\frac{\pi_p}{n_p} { \sum\limits_{i=1}^{n_p} } l(g(x_i^{p}),-1))
    \label{nnpu}
    \end{aligned}
\end{equation}

where $\pi$ is the positive class prior, $l(\cdot, \cdot)$ is any surrogate loss of zero-one loss \cite{du2015convex}, $n_p$ represents the number of labeled positive instances, $n_u$ represents the number of unlabeled instances, $x_i^p$ and $x_i^u$ denote instances in the labeled positive set and the unlabeled set, respectively. All cases shared the same domain model layer in training process. The details are formalized in Alg. \ref{pppu}.

\begin{algorithm}
\small
	\renewcommand{\algorithmicrequire}{\textbf{Input:}}
	\renewcommand{\algorithmicensure}{\textbf{Output:}}
	\caption{PURL Training}
	\label{purl}
	\begin{algorithmic}[1]
		\REQUIRE Training cases fact descriptions set $\overline{D}$; N-hop subgraphs set $G'$; positive class priors set $\pi$; training epochs $E$; step size $\eta$; discount factor $\gamma$; LLM
        \ENSURE $\text{Domain\_pu}$ model parameters $\theta$ and case-specific NeuralUCB parameters set $\{\phi_1, \phi_2, \ldots, \phi_{|X|}\}$
        \STATE Randomly initialize $\theta$ and $\{\phi_1, \phi_2, \ldots, \phi_{|X|}\}$ \

        \STATE $\theta$ = \text{Train\_Domain\_PU}($\overline{D}, G', \pi, E, \eta, \gamma$) using Alg. \ref{pppu}
        \STATE Calculate reward R based on Eq. \ref{loss}

        \FOR{$i = 1, 2, \ldots, {|X|}$}
        \FOR{$t = 1, 2, \ldots, T$}
            \STATE Observe $\left\{x_{t,F_i^j}\right\}_{F_i^j\in F_i}$
            \STATE Select action $F_i^j$ with the largest upper confidence bound using NeuralUCB \citep{zhou2020neural}
            \STATE Play action $\smash{F_i^j}$ and observe reward $R(F_i^j)$

            \STATE Update NeuralUCB parameters $\phi_i$ using $\left\{x_{z,\left(F_i^{ }\right)_z}\right\}_{z\ =\ 1}^t$ and $\left\{R_{z,\left(F_i^{ }\right)_z}\right\}_{z\ =\ 1}^t$
        \ENDFOR\
        \ENDFOR\
		\STATE \textbf{Return} $\text{Domain\_pu}$ model with updated parameters $\theta$ and case-specific NeuralUCB with updated parameters set $\{\phi_1, \phi_2, \ldots, \phi_{|X|}\}$
	\end{algorithmic}  
\end{algorithm}

\subsubsection{LLM Reading Comprehension}
For each masked case description \(\overline{D_i}\) and its corresponding candidate facts set \(F_i\), we utilize the reading comprehension capabilities of the LLM to assess the significance of each candidate fact \(F_i^j \in F_i\). Incorporating \(F_i^j\) into the masked description aims to enrich the case narrative by reconstructing the fact description and generating an enhanced court view \(\overline{C_i(+F_i^j)}\). This enhanced view, \(\overline{C_i(+F_i^j)}\), is then used to answer the questions set \(Q_i\), derived from the original court views text \(C_i\). We evaluate the reading comprehension score of $F_i^j$ for case $\overline{D_i}$ as:

\begin{equation}
\small
\begin{aligned}
    RC_{Score}\left(\overline{D_i},F_i^j\ ;\ LLM\right)\ =\frac{\left|S\left(\overline{C_i+F_i^j},\ Q_i\right)\right|}{\left|Q_i\right|}
\end{aligned}
    \label{lc_s}
\end{equation}
where $S\left(\overline{C_i+F_i^j},\ Q_i\right)$ is a subset of $Q_i$ that consists of questions correctly answered using  \(\overline{C_i(+F_i^j)}\).

\subsubsection{Model Training}
For each case description text $\overline{D_i}$ and candidate fact nodes $F_i$, we denote their corresponding embeddings are $h_i$ and $H_i$ ($H_i$ is matrix of all fact nodes embeddings). We calculate two types of rewards: 1)$R_{pu}$, the probability score from the domain-specific PU model, calculated as $P\left(h_i,\ H_i^j\ ;\ \theta\right)$, where $\theta$ is the domain positive-unlabeled model; 2) $R_{lmrc}$, the language model reading comprehension score calculated as Eq. \ref{lc_s}. To optimize fact node selection, we employ reinforcement learning (bandit), which allows the model to refine its choice dynamically. The selected fact node is then formulated as the next query for the legal client. The overall reward function is:
\begin{equation}
    \begin{aligned}
        R_{total}\ =\ R_{pu}\ +\ \lambda R_{lmrc}
    \end{aligned}
    \label{loss}
\end{equation}

where $\lambda$ adjusts the contribution of the LMRC score to the total reward, and we set 0.5 in our model. By leveraging bandit learning for its flexibility, our model surpasses traditional methods by better handling uncertainty in selecting the most relevant fact node for next question generation.





\subsubsection{Model Inference}
During inference, the process involves mapping the provided case description text to its corresponding fact-rule graph. From this graph, a subgraph is extracted from the overarching knowledge graph, which yields a set of candidate facts. For each of these facts, the corresponding bandit algorithm is employed to facilitate the selection process.

\section{Experiments}

\subsection{Data Construction}
Addressing the notable absence of English-language datasets for court views generation (CVG), our research introduces the first US-CVG dataset, derived from criminal legal documents \citep{caselaw2024}. Unlike the structured format of Chinese legal documents, U.S. case law's complexity and length (averaging over 7694 tokens) pose significant challenges for CVG tasks, often surpassing the token limits of advanced LLMs.

Our dataset leverages the IRAC framework to process US caselaw documents, using GPT-4.0 to summarize and extract fact descriptions and court views, which enables the reduction of the total token length to a structured IRAC format. Additionally, we create fact-rule graphs for each case to support complex analysis. The dataset's integrity is ensured through meticulous review by three legal professionals, after which it is segmented into training, validation, and test sets in an 8:1:1 ratio.

This pioneering dataset fills a critical void for English-language CVG research, offering a structured approach to legal document analysis that paves the way for future computational legal studies.
\begin{table}[htp]
\small
  \centering
  \caption{Statistics of US legal dataset.}
  \setlength{\tabcolsep}{3mm}{   
    \begin{tabular}{cc}
    \hline
     Type & US  \\
    \hline
    $\# \text{Train}$ & 19491  \\
    $\# \text{Dev}$ & 2436 \\
    $\# \text{Test}$ & 2437  \\
    $\text{Avg.}\#$\text{tokens in fact desc.} & 264 \\
    $ \text{Avg.}\#$\text{tokens in court views} & 138\\
    \hline
    \end{tabular}%
    }
  \label{data_overview}%
\end{table}%

\subsection{Experimental Settings}
The experimental setup for our D3LM model utilized the Llama2-13B as the base architecture. The sampling parameters were set with a temperature of 0.8 and a top\_p value for nucleus sampling at 0.9, alongside a maximum token count of 4096. We initialized word embeddings with RoBERTa, while node embeddings were randomly generated. The model's Positive-Unlabeled learning component consisted of six Multi-Layer Perceptron (MLP) layers. The learning rate for the PURL was established at $1e^{-4}$, paired with a batch size of 2000. Conducted on an NVIDIA A100 GPU, our experiments spanned over 100 epochs to ensure comprehensive training.
\begin{table*}[htp]
\small
  \centering
  \caption{Results on US court views generation. \textbf{\best{Best}} and \textbf{\secondb{second}} values are both highlighted.}
  \label{auto_res}
  \setlength{\tabcolsep}{10pt} 
  \renewcommand{\arraystretch}{0.9}
  \begin{tabular}{lcccccc}
    \toprule
    \multirow{2}{*}{Method} & \multicolumn{3}{c}{ROUGE(\%)} & \multicolumn{3}{c}{BLEU(\%)} \\
    \cmidrule(lr){2-4} \cmidrule(lr){5-7}
    & R-1 & R-2 & R-L & B-1 & B-2 & B-N \\
    \midrule
    BM25 & 30.8 & 16.1 & 28.2 & 20.1 & 13.9 & 10.9 \\
    AC-NLG & 35.8 & 18.5 & 32.7 & 23.6 & 16.2 & 12.7 \\
    Llama2-13B-ST & 37.5 & 24.3 & 37.5 & 20.8 & 14.6 & 11.4 \\
    FT-Llama2-13B-ST & 39.7 & 26.5 & 41.8 & 24.3 & 17.4 & 13.2 \\
    GPT4.0-ST & \textbf{\secondb{60.3}} & \textbf{\secondb{42.7}} & \textbf{\secondb{57.1}} & 27.2 & 18.2 & 14.1 \\
    D3LMw/oPU & 51.5 & 31.3 & 51.5 & 35.1 & 25.8 & 21.5 \\
    D3LMw/oLMRC & 50.0 & 27.9 & 44.1 & 35.6 & 27.4 & 22.7 \\
    D3LMw/oRF & 54.1 & 31.9 & 54.1 & \textbf{\best{39.6}} & \textbf{\secondb{29.9}} & \textbf{\secondb{25.1}} \\
    \midrule
    D3LM & \textbf{\best{63.3}} & \textbf{\best{53.1}} & \textbf{\best{59.2}} & \textbf{\secondb{38.7}} & \textbf{\best{31.7}} & \textbf{\best{26.9}} \\
    \bottomrule
  \end{tabular}
\end{table*}

\subsection{Comparisons with Baselines}
\textbf{Evaluation Metrics.} To evaluate our model, we implemented both automatic and human assessment methods. For automatic evaluation, we used Rouge (\textbf{Rouge-1}, \textbf{Rouge-2}, and \textbf{Rouge-L} for LCS-based statistics) and BLEU scores (\textbf{BLEU-1}, \textbf{BLEU-2}, and an average of BLEU-1 to BLEU-4, denoted as \textbf{BLEU-N}), which are benchmarks in natural language generation tasks. Furthermore, human judgment focused on three aspects: 1) \textbf{fluency}, 2) \textbf{accuracy}, and 3) \textbf{adoptable} of rationales in generating court views, rated on a five-point scale (with 5 being the highest). This evaluation was conducted by three legal professionals, ensuring a comprehensive and expert assessment of the model's performance in producing contextually accurate and legally coherent text. This streamlined approach balances thoroughness with brevity, maintaining the integrity of the evaluation process while ensuring clarity and conciseness.\\
\begin{table}[ht]
\small
  \centering
  \caption{Results of human judgement}
  \label{hum_res}
  \setlength{\tabcolsep}{5pt} 
  \begin{tabular}{lccc}
    \toprule
    MODEL & FLUENT & ACC. &ADOPT \\
    \midrule
    Llama2-13B-ST & 4.68 & 4.32 & 4.03 \\
    GPT4.0-ST & \textbf{\best{4.74}} & \textbf{\secondb{4.39}} & \textbf{\secondb{4.06}} \\
    \midrule
    D3LM & \textbf{\secondb{4.71}} & \textbf{\best{4.48}} & \textbf{\best{4.19}} \\
    \bottomrule
  \end{tabular}
\end{table} \\
\begin{figure*}[t]
  \centering
  \includegraphics[width=1.0\linewidth]{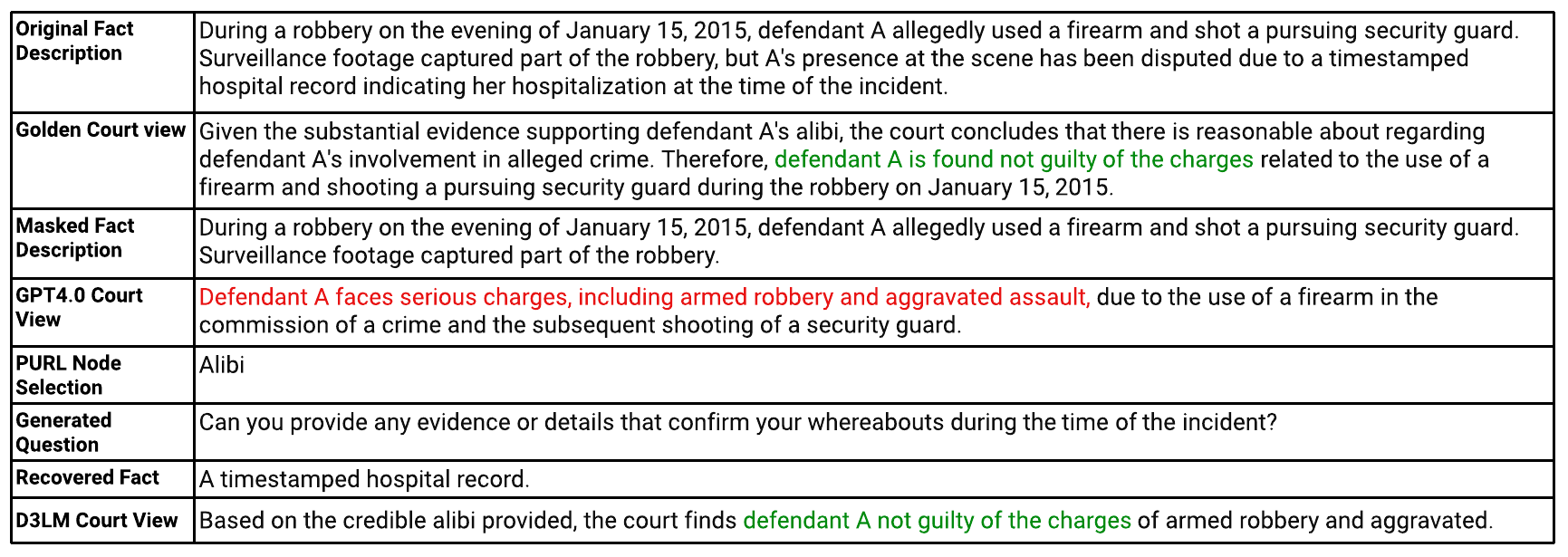}
  \caption{Case Study.}
  \label{caseaaa}
\end{figure*}
\textbf{Baselines.} We implement the following baselines for comparison:\\
$\bullet$ \textbf{BM25} \citep{robertson1994some} is a classic retrieval baseline that identifies the most closely related text content between case fact descriptions and court views' rationale sections. In our study, we treat all court views data (train, dev, test) as a database, retrieving the court views with the highest relevance score to each case's fact description using BM25 as our baseline model.\\
$\bullet$ \textbf{AC-NLG} \citep{wu2020biased}is an attention and counterfactual-based generative model, designed to transform fact descriptions into court views by training with these elements as input and output, respectively.  \\
$\bullet$ \textbf{Llama2-13B-ST} \citep{touvron2023llama} is a LLM developed by Meta, which utilizes prompts and fact descriptions as inputs to generate court views responses. (ST means single-turn interaction between user and LLM.)\\
$\bullet$ \textbf{FT-Llama2-13B-ST} is a fine-tuned version of Llama2-13B-ST. We use the training data to fine-tune the model, employing fact descriptions as input and court views as output in the fine-tune process.\\
$\bullet$ \textbf{GPT4.0-ST} \citep{achiam2023gpt}is an LLM created by OpenAI, employing prompts and fact descriptions to interactively produce court views responses. \\
$\bullet$ \textbf{D3LMw/oPU} is a variant of our proposed D3LM model that does not incorporate positive-unlabeled learning in its framework.\\
$\bullet$ \textbf{D3LMw/oLMRC} is a variant of our proposed D3LM model excluding the use of language model reading comprehension.\\
$\bullet$ \textbf{D3LMw/oRF} is a variant of our proposed D3LM model that operates without the integration of reinforcement learning. \\
\textbf{Experimental Result.} 
In the automatic evaluation detailed in Tab. \ref{auto_res}, the D3LM model showcases superior performance in generating U.S. court views, achieving the highest ROUGE and BLEU scores. Notably, GPT4.0-ST stands out as a particularly strong competitor, especially in terms of ROUGE metrics. Furthermore, the variant D3LMw/oRF registers impressively high BLEU scores, highlighting its adeptness at capturing the subtle linguistic qualities of the generated court views. The comparison between D3LM and its ablated versions underlines the significance of each individual module, confirming that every element plays a vital role in the overall success of the model in the complex task of legal text generation.

In the human evaluation, as illustrated by Tab. \ref{hum_res}, our D3LM model demonstrates superior performance in accuracy and adaptability, achieving scores of 4.48 and 4.19, respectively, which marginally exceed those of GPT4.0-ST by about 0.1 points in each metric. GPT4.0-ST, however, takes the lead in fluency with a score of 4.74, proving to be a formidable competitor in semantic coherence. This could be attributed, in part, to our model's foundation on the Llama2-13B architecture. Despite this, the results emphasize D3LM's capability to generate precise and contextually adaptable court views, underlining its efficacy in comparison to a notably strong rival in legal text generation.


\begin{table}[ht]
\small
  \centering
  \caption{Results of usability testing}
  \label{usability}
  \setlength{\tabcolsep}{5pt} 
  \begin{tabular}{lccc}
    \toprule
    Model & Reli. & Sati. & Pref.(\%) \\
    \midrule
    GPT4.0-ST & \secondb{3.03} & \secondb{3.17} & \secondb{37.7} \\
    D3LM & \textbf{\best{3.58}} & \textbf{\best{3.65}} & \textbf{\best{62.3}} \\
    \bottomrule
  \end{tabular}
\end{table}
\subsection{Human Acceptability}
To ensure fairness when comparing our model's metrics with those of traditional LLMs, we conducted a usability testing to gauge user experience and perceptions regarding our approach, which involves greater interaction with legal clients and thus, allows our model to gather more case-specific information. We invited 69 participants to evaluate their experiences and opinions on the state-of-the-art LLM, GPT-4.0, versus our model, D3LM, across three key metrics: 1) \textbf{Reliability} (Reli.), the perceived reliability of the models; 2) \textbf{Satisfaction} (Sati.), the level of satisfaction with the models; and 3) \textbf{Preference} (Pref.), their preference for either model. We employed a five-point scale for both reliability and satisfaction assessments (where 5 represents the highest rating), and participants were asked to indicate a preference between GPT-4.0 and D3LM.

The findings, detailed in Table \ref{usability}, revealed that on legal tasks, users reported approximately 0.5 points higher reliability and satisfaction scores for D3LM compared to GPT-4.0. Moreover, 62.3\% of participants expressed a preference for using D3LM over GPT-4.0. These results underscore the acceptance and appreciation for our model's interactive questioning approach, highlighting D3LM's enhanced accuracy and reliability in legal applications.

\subsection{Case Study}
Fig. \ref{caseaaa} presents a comparison of court views generated by GPT4.0 and the proposed D3LM method for a masked fact description in a specific US criminal case. Upon comparison with the actual court's view, it is evident that our PURL feature within D3LM can precisely recover the fact "Alibi" and systematically generate questions to elicit further information about the case. Consequently, D3LM produces a more accurate court view than GPT4.0.

\section{Conclusion and Future Work}


Our study introduces D3LM, an innovative approach in legal assistant systems for court view generation, diverging from traditional methods by simulating real lawyer-client interactions. By innovatively employing Diagnostic Analysis through continuous user engagement, we enable the generation of more accurate and case-specific court views. The introduction of our PURL algorithm significantly enhances the precision of information extraction. Additionally, we contribute a novel English-language dataset for American legal cases, meticulously validated by legal professionals, addressing the scarcity in legal analysis datasets and setting new benchmarks in legal informatics. Future work will explore extending this model to other domains, such as medical and consultancy, leveraging its capabilities for broader application.

\section*{Limitations}

In our study, we explored the potential of Large Language Models (LLMs) to improve legal assistance through active engagement with clients, prompting for case-specific details. However, our findings underscore several limitations: the PURL algorithm's effectiveness is confined to the criminal cases domain due to its reliance on a specialized knowledge graph, which is not only resource-intensive to create but also limits cross-domain applicability. Moreover, our evaluation was restricted to English language cases, overlooking the model's performance in diverse linguistic contexts-a critical consideration for global legal system applicability. Furthermore, the model demands significant computational and human annotation resources, with its operational speed lagging behind that of existing large models, potentially hindering its practicality in time-sensitive settings. These challenges underscore the need for future research to focus on optimizing computational efficiency, broadening linguistic and domain applicability, and reducing resource dependencies, aiming to enhance the practicality and accessibility of LLMs in legal assistance across a wider range of contexts.

\section*{Acknowledgements}

\bibliography{anthology,custom}
\bibliographystyle{acl_natbib}

\nocite{}

\appendix

\section{Appendix}
\label{sec:appendix}

\subsection{node attention}
\label{node_atta}

Given a masked case description embedding $h_i$ and its corresponding subgraph node embeddings $H_i$. We conduct node embedding as following:
\begin{equation}
\small
    \begin{aligned}
        e_{ij}\ =\ \sigma\left(W\cdot\left[h_i\ ,H_i^j\right]\ \ +b\right)
     \label{att}
    \end{aligned}
\end{equation}


\begin{equation}
\small
    \begin{aligned}
            \alpha_{ij}\ =
    \ \frac{\exp\left(\sigma\left(W\cdot\left[h_i\ ,H_i^j\right]\ \ +b\right)\right)}{\sum_{k\in H_i}^{ }\exp\left(\sigma\left(W\cdot\left[h_i\ ,H_i^k\right]\ \ +b\right)\right)}
    \label{att2}
    \end{aligned}
\end{equation}

\begin{equation}
\small
    \begin{aligned}
        Z_i\ =\sum _{j\in H_i}^{ }\alpha _{ij}h_{ij}
    \label{sum}
    \end{aligned}
\end{equation}
where $\sigma$ denotes $Relu$ activation function, $\left[.,.\right]$ represents embedding concatenation. Furthermore, we concatenate the $Z_i$ with each fact node embedding $H_i^j$, and pass through a MLP to obtain output probability as:
\begin{equation}
\small
    \begin{aligned}
        P\left(h_i,H_i^j\right)\ \ =\ \phi \left(\left[Z_i,\ H_i^j\right]\right)
    \label{pro}
    \end{aligned}
\end{equation}




\subsection{Prompt details}
\label{prompts}
\subsubsection{Case type classification}
"Please select the most appropriate category from the following options: Contract Law, Criminal Law, Property Law, Intellectual Property Law, Business Law, Tax Law, Estate and Trust Law, Family Law, Administrative Law, Civil Law, Tort Law, Bankruptcy Law, Environmental Law. Provide the category name that best fits the case described below."
\subsubsection{IRAC summarization}
"Based on the case description, please generate legal case IRAC (Issue, Rule, Analysis, Conclusion) in following pattern: \\
    **Issue:** \\
    Identifying the legal question or questions that need to be addressed in the legal case.\\
    **Rule:** \\
    Stating the applicable legal rules, statutes, regulations, or case law that pertain to the identified issue.\\
    **Analysis:** \\
    Applying the relevant legal rules to the facts or circumstances of the case. This involves explaining how the rules should be interpreted and applied.\\
    **Conclusion:** \\
    Summarizing the court judgement to the legal issue."
\subsubsection{Fact-Rule graph extraction}
"I'm developing a project to analyze legal cases deeply by distilling case texts into concise formats for graphical representation. Help me extract key facts influencing court decisions, the applicable rules, and the relationships between them without considering the decision's outcome. Facts should be generalized (e.g., "Contract", "Damages") and rules brief (e.g., "Civil Law", "Contract Law"). Edges should either show a fact's dependency on another fact ("Depends On") or its compliance with or violation of a rule ("Complies With"/"Violates"). Provide three Python lists for networkx: facts, rules, and edges, with edges formatted as ('Fact', 'Rule', {'relation': 'Relation Type'}). Ensure edges accurately reflect the connections without adding non-existent links. Respond in English with only the lists in a single Python chunk, without extra text or explanation."
\subsubsection{Case reconstruction}
"I have a detailed description of a legal case, including the Issue, Rule, and Analysis of the case. I would like you to help me rewrite this case's IRA, but please disregard the specific aspects that I will mention below."
\subsubsection{LLM reading comprehension}
Question generation:\\
"I will provide a conclusion of a legal case, please generate 10 unique questions and corresponding answers based on the conclusion information. Please do not give me explanation of the question, greeting words and any annotation words, just give me generated questions."

Scoring:\\
"I will provide a conclusion of a legal case, and 10 unique questions and golden answers, please use the given conclusion to answer the 10 questions, if the conclusion can correctly answer one question, it will get one score, the maximum score is 10, please tell me the final score of the conclusion:"

\subsection{Comparison between PU and PN}
To verify the effectiveness of our positive-unlabel module, we conducted a comparative experiment between the PU (Positive-Unlabeled) approach and the PN (Positive-Negative) approach. In the PN setting, all unlabeled candidate nodes are treated as negative. The results of the experiment are presented in Table \ref{PU_vs_PN}. The accuracy, recall, precision, F1, F2, and AUC scores for PU are significantly higher than those for PN.
\begin{table}[ht]
\small
  \centering
  \caption{Positive-Unalbeled vs Positive-Negative}
  \label{pupn}
  \setlength{\tabcolsep}{5pt} 
  \begin{tabular}{lcccccc}
    \toprule
    Model & Acc. & Rec. & Pre. &F1 & F2 &AUC \\
    \midrule
    PN & \secondb{65.4} & \secondb{54.9} & \secondb{78.8} &\secondb{42.3} & \secondb{51.1} & \secondb{90.9}\\
    PU & \textbf{\best{87.3}} & \textbf{\best{86.1}} & \textbf{\best{86.8}} &\textbf{\best{86.4}} & \textbf{\best{86.3}} & \textbf{\best{92.8}}\\
    \bottomrule
  \end{tabular}
  \label{PU_vs_PN}
\end{table}

\subsection{CVG dataset generation algorithm}
\label{CVG_algorithm_pseudo_code}
\begin{algorithm}
\small
	\renewcommand{\algorithmicrequire}{\textbf{Input:}}
	\renewcommand{\algorithmicensure}{\textbf{Output:}}
	\caption{CVG Dataset Generation via LLM}
	\label{cvg}
	\begin{algorithmic}[1]
		\REQUIRE Legal cases set $X$
        \ENSURE CVG dataset $\{D, C, \overline{D}, Q, F,\hat{G}, G', \overline{G}, $ $ G\}$  \

        \STATE Initialize $D, C, \overline{D}, Q, F, \hat{G}, G', \overline{G}, G$ are all as empty lists\
        \FOR{$i = 1, 2, \ldots, {|X|}$}
            \STATE $G_{i} = LLM_{\text{extract}}(X_i)$
            \STATE $D_i\ ,\ C_i=LLM_{\text{summarize}}(X_i)$
            \STATE $Q_i=LLM_{\text{generate}}(C_i)$
            \STATE $\overline{G_i}= Masked(G_i)$
            \STATE $\overline{D_i}=LLM_{\text{reconstruct}}(D_i)$
            \STATE $D\leftarrow D + D_i$   \qquad $C\leftarrow C + C_i$\\
             $\overline{D}\leftarrow\overline{D_i} + D_i$ \qquad$Q\leftarrow Q+ Q_i$\\
             $\hat{G}\leftarrow \hat{G}+ G_i$ \qquad\hspace{0.1cm}$\overline{G_i}\leftarrow \overline{G}+ \overline{G_i}$\

        \ENDFOR\
        \STATE $G = Merge(\hat{G})$
        \FOR{$i = 1, 2, \ldots, {|X|}$}
            \STATE $G_i' = G_{\text{N-hop}}(\overline{G_i})$
            \STATE $F_i=Extract(G_i')$
            \STATE $G'\leftarrow G' + G_i'$ \qquad $F \leftarrow F+ F_i$\
        \ENDFOR\
        \STATE Human\_Check $\{D, C, \overline{D}, Q, F, \hat{G}, G', \overline{G},  G\}$ \

		\STATE \textbf{Return} LLM-Generated CVG dataset $\{D, C, \overline{D}, Q, F, \hat{G}, G' , \overline{G},  G\}$  \
	\end{algorithmic}  
\end{algorithm}

\vspace{50mm}
\subsection{Domain PU model training}
\label{Domain_PU_training_pseudo_code}\textit{}
\begin{algorithm}
\small
	\renewcommand{\algorithmicrequire}{\textbf{Input:}}
	\renewcommand{\algorithmicensure}{\textbf{Output:}}
	\caption{\text{Train\_Domain\_PU}}
	\label{pppu}
	\begin{algorithmic}[1]
		\REQUIRE Training cases fact descriptions set $\overline{D}$; N-hop subgraphs set $G'$; positive class priors set $\pi$; training epochs $E$; step size $\eta$; discount factor $\gamma$
        \ENSURE $\text{Domain\_pu}$ model parameters $\theta$ 
        \STATE Randomly initialize $\theta$ \
        \STATE Let A be an external SGD-like stochastic optimization algorithm such as Adam (reference)
        
        \FOR{$e = 1, 2, \ldots, E$}
        \FOR{$i = 1, 2, \ldots, {|\overline{D}|}$}
            \STATE $h_{i} = Roberta(\overline{D_i})$
            \STATE $H_{i} = DiGCN(G'_i)$
            \STATE Denote positive instances in $H_{i}$ as $x_p^i$, unlabeled instances in $H_{i}$ as $x_u^i$
            \IF{$\widehat{R}_{u}^-\left(g;\ x_u^i\right)-\pi_p\widehat{R}_{p}^-\left(g;\ x_p^i\right)\ge0$}
                \STATE Set gradient $\nabla_{\theta}\widehat{R}_{pu}\left(g;\ x_p^i,x_u^i\right)$
                \STATE Update $\theta$ by A with current step size $\eta$
            \ELSE
                \STATE Set gradient $\nabla_{\theta}(\pi_p\widehat{R}_{p}^-\left(g;\ x_p^i\right)-\widehat{R}_{u}^-\left(g;\ x_u^i\right))$
                \STATE Update $\theta$ by A with discounted step size $\gamma\eta$
            \ENDIF
        \ENDFOR\
        \ENDFOR\
		\STATE \textbf{Return} $\text{Domain\_pu}$ model with updated parameters $\theta$
	\end{algorithmic}  
\end{algorithm}

\begin{figure*}[t]
  \centering
  \includegraphics[width=1.0\linewidth]{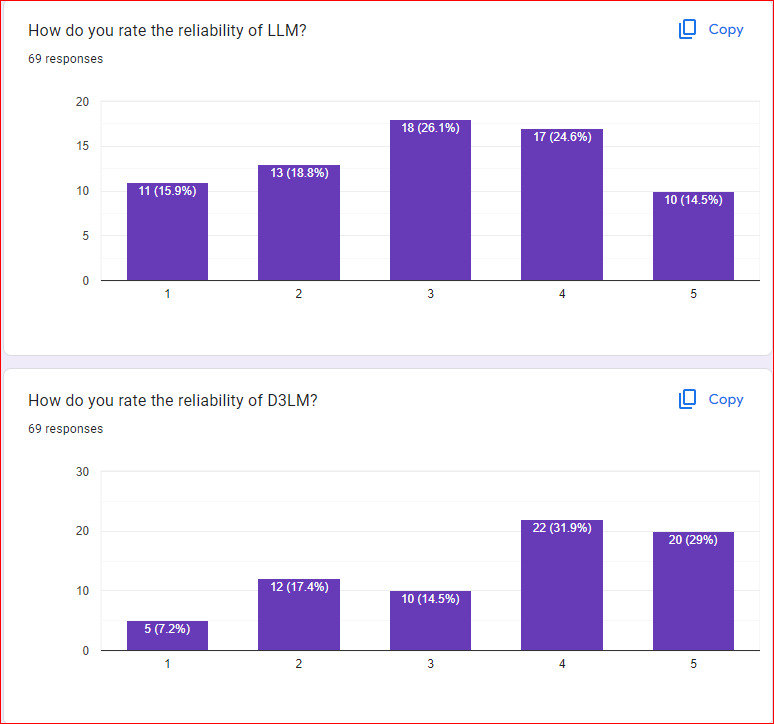}
  \caption{Reliability result}
  \label{reli}
\end{figure*}

\begin{figure*}[t]
  \centering
  \includegraphics[width=1.0\linewidth]{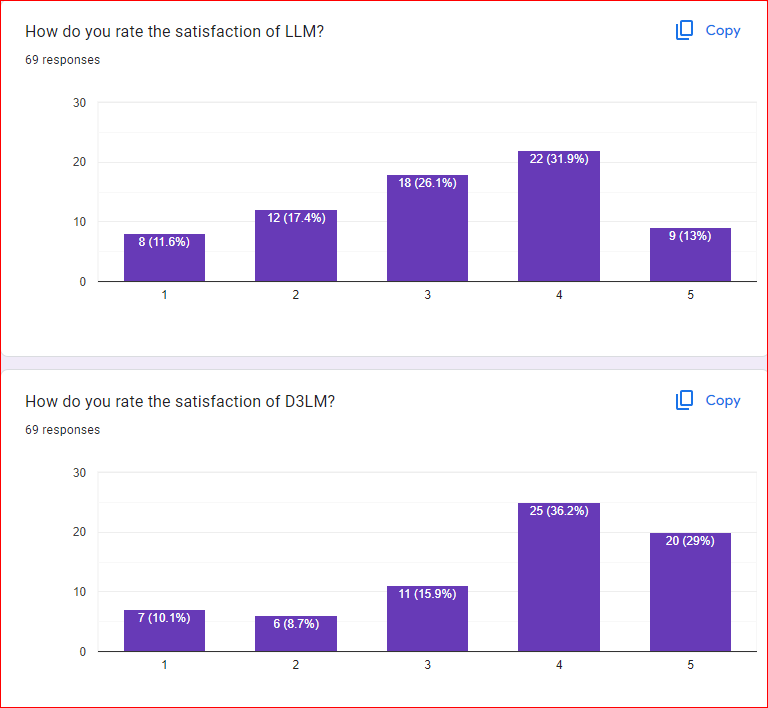}
  \caption{Satisfaction result}
  \label{reli}
\end{figure*}
\begin{figure*}[t]
  \centering
  \includegraphics[width=1.0\linewidth]{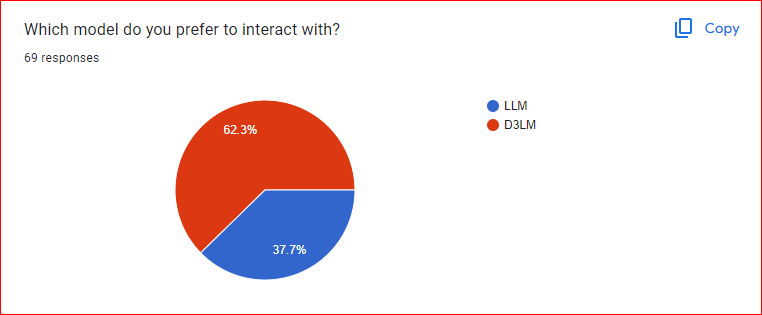}
  \caption{Preference result}
  \label{reli}
\end{figure*}

\end{document}